\documentclass{article}
\usepackage{spconf,amsmath,graphicx}
\usepackage{xcolor}
\usepackage{booktabs}
\usepackage{multirow}
\usepackage{array}
\usepackage{colortbl}
\usepackage{subcaption}

\title{Handling Multiple Hypotheses in Coarse-to-Fine Dense Image Matching}
\name{$^1$Matthieu Vilain, $^1$Remi Giraud, $^1$Yannick Berthoumieu, and $^1$Guillaume Bourmaud}
\address{$^1$Univ. Bordeaux, CNRS, Bordeaux INP, IMS, UMR 5218, F-33400 Talence, France}

\definecolor{gray}{rgb}{0.5,0.5,0.5}
\definecolor{orange}{rgb}{1.0,0.5,0}
\definecolor{green}{rgb}{0.1,0.8,0.1}
\definecolor{brown}{rgb}{1.0,0.,1.0}
\definecolor{black}{rgb}{0.0,0.0,0}
\definecolor{red}{rgb}{1.0,0.1,0.1}
\definecolor{blue}{rgb}{0.1,0.1,1.0}

\newcommand{\green}[1]{\textcolor{green}{#1}}

\newcommand{\red}[1]{\textcolor{red}{#1}}

\def\m#1{\ensuremath{\mathtt{#1}}}

\def\vec#1{\ensuremath{\mathbf{#1}}}

\def\mI{\m I}
\def\mF{\m F}

\def\mC{\m C}

\def\vp{\vec p}

\def\parr#1{\left(#1\right)}
\def\curl#1{\left\{#1\right\}}
\def\nearest#1{\left\lceil  #1\right\rfloor}

\def\tr{^{\top}}

\begin{document}
%
\maketitle
\begin{abstract}
{Dense image matching aims to find a correspondent for every pixel of a source image in a partially overlapping target image.
State-of-the-art methods typically rely on a coarse-to-fine mechanism where a single correspondent hypothesis is produced per source location at each scale. 
In challenging cases -- such as at depth discontinuities or when the target image is a strong zoom-in of the source image -- the correspondents of neighboring source locations are often widely spread and predicting a single correspondent hypothesis per source location at each scale may lead to erroneous matches.
In this paper, we investigate the idea of predicting \emph{multiple} correspondent hypotheses per source location at each scale instead.
We consider a beam search strategy to propagate multiple hypotheses at each scale and propose integrating these multiple hypotheses into cross-attention layers, resulting in a novel dense matching architecture called BEAMER.
BEAMER learns to preserve and propagate multiple hypotheses across scales, making it significantly more robust than state-of-the-art methods, especially at depth discontinuities or when the target image is a strong zoom-in of the source image.
}
\end{abstract}

\section{Introduction}
\label{sec:intro}

\begin{figure}[ht]
    \centering
    \rotatebox{90}{\hspace{0.7cm} Target \hspace{1.4cm} Source \hspace{0.cm} }
    \includegraphics[width=0.90\linewidth]{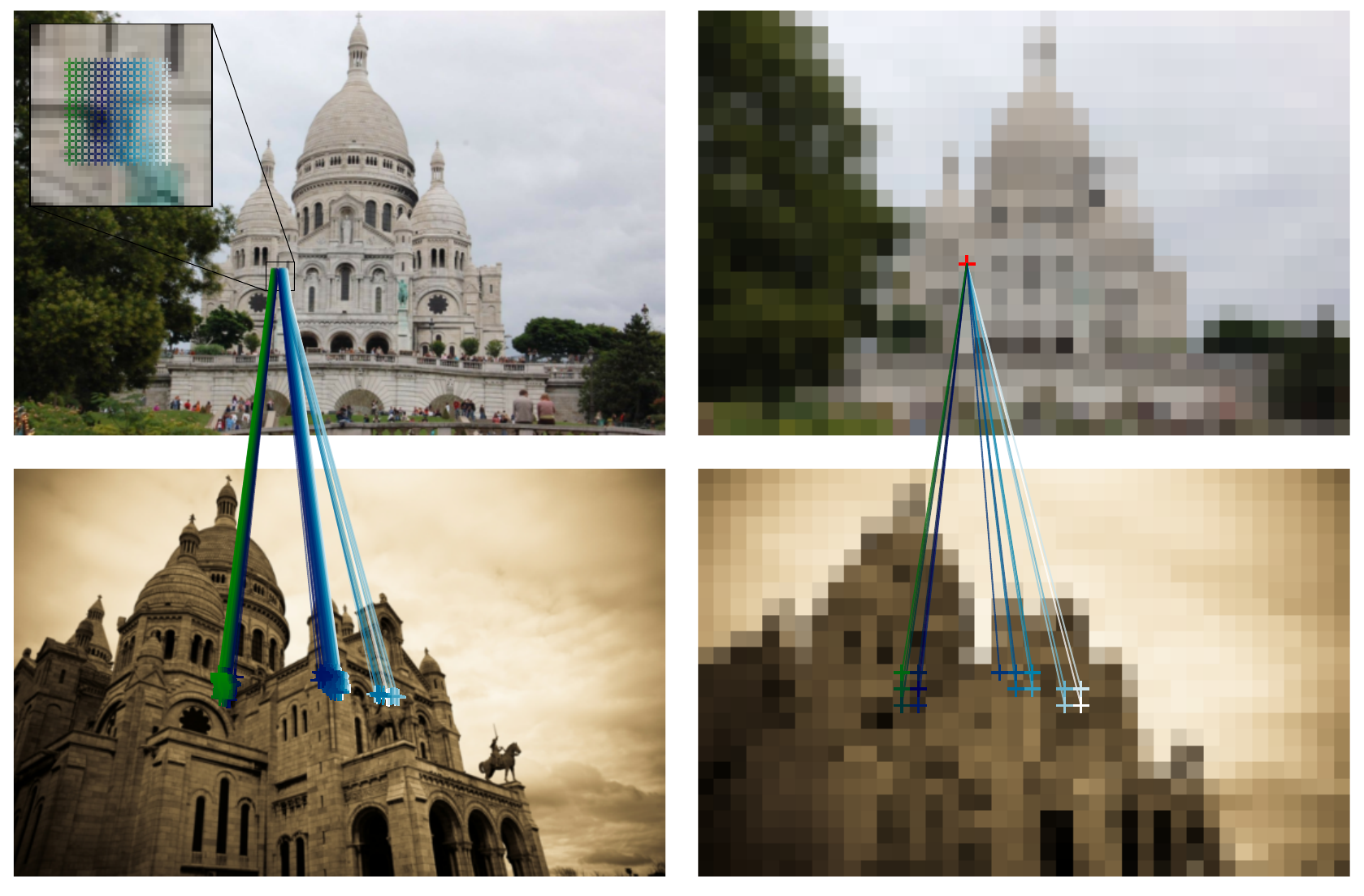}\\[-1.5ex]
    {\footnotesize
    \begin{minipage}{\columnwidth}
    \centering
    \hspace{0.05\columnwidth}
        \begin{minipage}{0.44\columnwidth}
        \centering
            GT correspondents at scale 1
        \end{minipage}
        \begin{minipage}{0.44\columnwidth}
        \centering
            GT correspondents at scale 1/16
        \end{minipage} 
    \end{minipage}
    \vspace{-4mm}}
    \caption{{\textbf{(Left)} Source locations in the neighborhood of a depth discontinuity (top-left: 16x16 patch) have ground truth (GT) correspondents in the target image that are widely spread. Here, the GT correspondents (bottom-left) are spread across three different modes in the target image. \textbf{(Right)} At scale 1/16, the source location $\textcolor{red}{+}$ (top-right), that corresponds to the 16x16 patch (top-left) at scale 1, has 15 GT correspondents that are widely spread (bottom-right). Thus, state-of-the-art dense image matching methods that rely on a coarse-to-fine mechanism and predict a single correspondent hypothesis per source location at each scale, have difficulties correctly establishing correspondences at depth discontinuities.}
 }
    \label{fig:teaser}
    \vspace{-5mm}
\end{figure}

\begin{figure*}[t]
\centering
{\footnotesize
\begin{tabular}{cccc}
$l=5$ \text{ } (scale 1/16) & $l=4$  \text{ } (scale 1/8) & $l=3$ \text{ } (scale 1/4) & $l=2$ \text{ } (scale 1/2) \\
\includegraphics[width=0.22\textwidth]{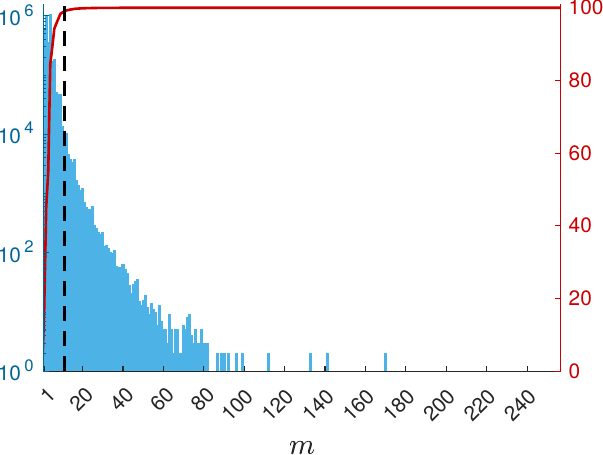}&
\includegraphics[width=0.22\textwidth]{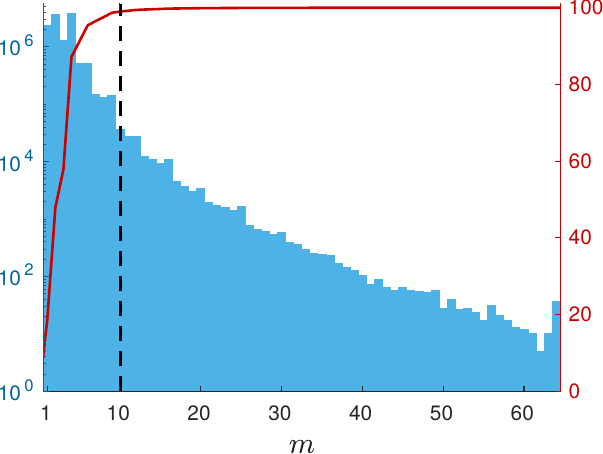}&
\includegraphics[width=0.22\textwidth]{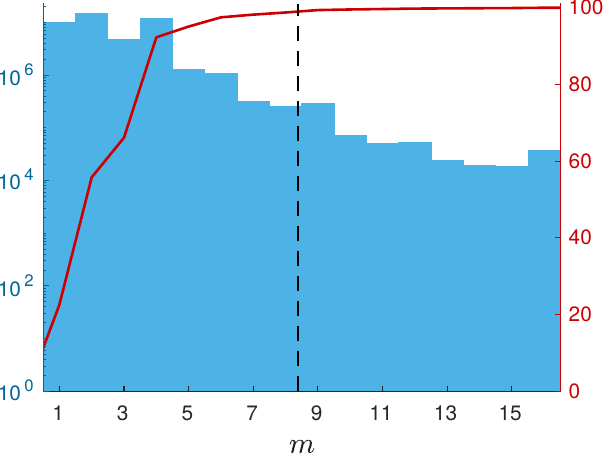}&
\includegraphics[width=0.22\textwidth]{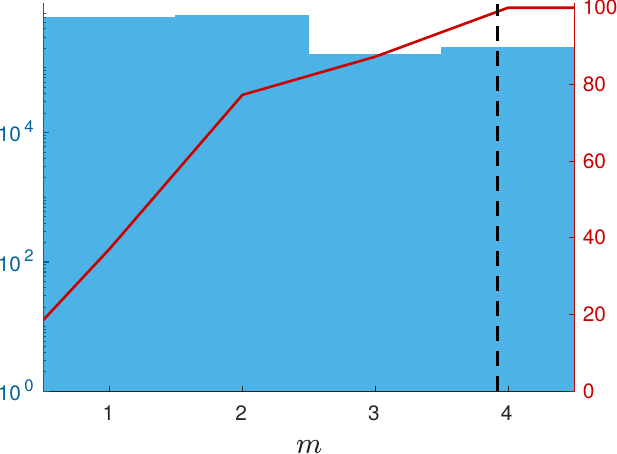}\\
\end{tabular}
\vspace{-3mm}
}
\caption{{\textbf{Experimental multiple hypothesis analysis on MegaDepth training samples.} For each coarse-to-fine scale, the histogram indicates the number of times a one-to-$m$ correspondence problem was found in the ground truth correspondences (left axis in log-scale). The red curve (right axis) is the cumulative histogram. The vertical dashed line marks the value of $m$ that encompasses 99\% of the distribution.}
}
\label{fig:exp_hypo_MD}
\end{figure*}

{Image matching seeks to establish correspondences between a pair of partially overlapping source and target images. This task is fundamental to various computer vision applications, including 3D reconstruction~\cite{heinly2015reconstructing}, simultaneous localization and mapping~\cite{sarlin2022lamar}, and visual localization~\cite{taira2018inloc}.\\}
{Image matching methods usually fall into one of the following three categories. \emph{Detector-based} methods~\cite{revaud2019r2d2,sarlin2020superglue,germain2020s2dnet,lindenberger2023lightglue,edstedt2024dedode,potje2024cvpr} establish correspondences between keypoints detected in the source image and in the target image.
\emph{Semi-dense methods} ~\cite{sun2021loftr,chen2022aspanforme,yu2023adaptive,cao2023improving,wang2024eloftr},
establish correspondences between source keypoints located on a coarse regular grid (hence the name "semi-dense") and the whole target image.
\emph{Dense methods}~\cite{truong2020glu,truong2021learning,edstedt2023dkm,truong2023pdc,zhu2023pmatch,edstedt2024roma}, try to find a correspondent for every pixel of the source image in the target image.}
{Dense methods currently outperform both detector-based methods and semi-dense methods on 
pose estimation benchmarks~\cite{taira2018inloc, berton2022deep, balntas2017hpatches, zhang2021reference}.}
{Therefore, in the rest of the paper, we focus on dense methods.}

{In this dense setting,
most methods rely on a \emph{coarse-to-fine} mechanism to efficiently search for the correspondent of each source pixel in the target image.}
{In practice,
at each scale of the coarse-to-fine mechanism, a \emph{single} correspondent hypothesis is produced per source location.}
{This approach
works well when there is no ambiguity, \emph{i.e.} when neighboring source locations at a given scale 
have target correspondents all within a small neighborhood.
}
{However, in challenging cases -- such as at depth discontinuities
or when the target image is a strong zoom-in of the source image -- the correspondents of neighboring source locations are often widely spread (see Fig.~\ref{fig:teaser}) and predicting a single correspondent hypothesis per source location at each scale may lead to erroneous matches, as acknowledged in the limitations of~\cite{edstedt2023dkm}.}
{In this paper, we investigate the idea of predicting \emph{multiple correspondent hypotheses} per source location at each scale.}

\noindent{Our contributions are as follows -
(i) We formulate dense matching as a series of coarse-to-fine classification steps, which leads us to consider a beam search strategy~\cite{furcy2005limited} for propagating multiple hypotheses at each scale.
(ii) We propose integrating these multiple hypotheses into cross-attention layers, resulting in a novel dense matching architecture, called BEAMER, which is able to learn to preserve and propagate multiple hypotheses across scales. 
(iii) Our experiments show that the performance of state-of-the-art 
dense matching methods significantly decreases when the correspondents of neighboring source locations are widely spread.
BEAMER is much more robust in these cases and significantly outperforms 
dense matching 
methods.}

\section{Method}

\begin{figure*}[t]
\centering
\vspace{-4mm}
\includegraphics[width=0.85\textwidth]{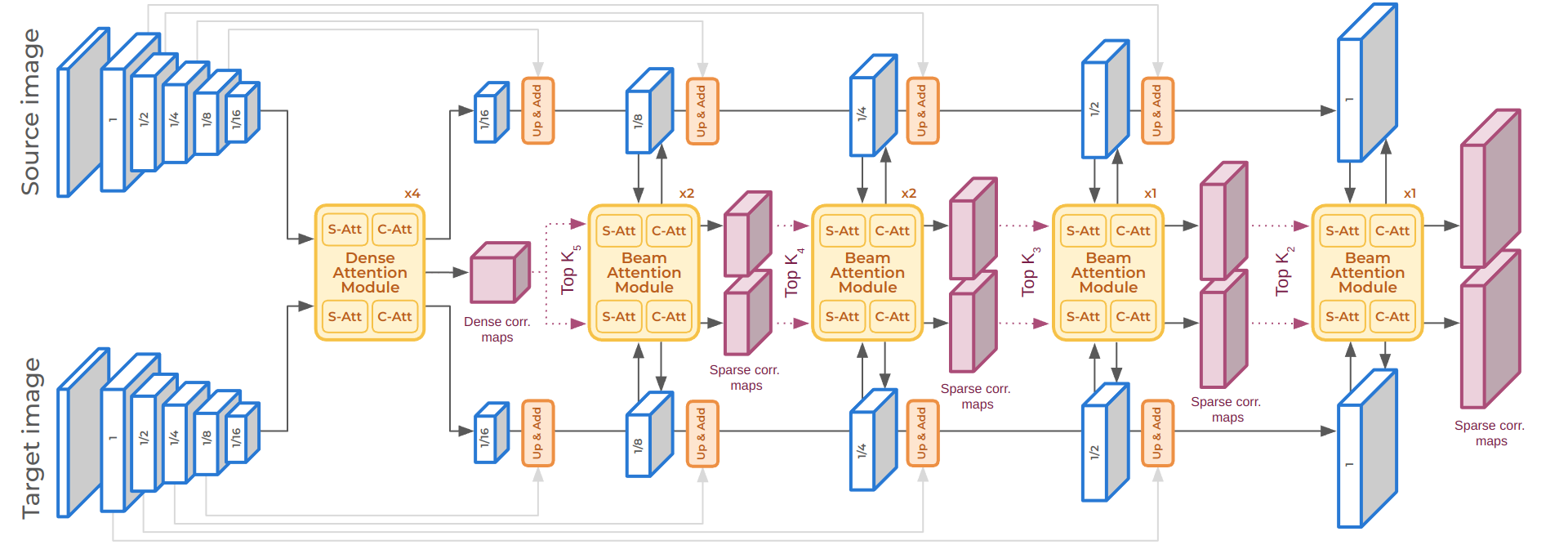}
\caption{\textbf{BEAMER architecture.} {Our novel architecture establishes dense correspondences bidirectionally, in a coarse-to-fine manner. At the coarsest scale, dense correspondence maps are computed that allow to initialize the beam search. Then at each scale, the beam-attention module employs the most promising correspondents from the previous correspondence maps in sparse cross-attention layers that let the features communicate with each other, and produce sparse correspondence maps. Sparse self-attention layers are also implemented. More details are available in the supplementary material.}
}
\label{fig:architecture}
\end{figure*}

\subsection{A series of coarse-to-fine classification steps}\label{sec:cf_dim_preliminaries}

{Dense image matching (DIM) aims to establish correspondences between a source image $\mI_\text{s}$ ($H_\text{s} \!\times\! W_\text{s} \!\times\! 3$) and a target image $\mI_\text{t}$ ($H_\text{t} \!\times\! W_\text{t} \!\times\! 3$). Given a neural network, we extract multi-scale dense feature maps for the source and target images, denoted as $\{\mF_\text{s}^l\}_{l=1...L}$ and $\{\mF_\text{t}^l\}_{l=1...L}$. At scale $l$, $\mF_\text{s}^l$ and $\mF_\text{t}^l$ are of size $\frac{H_\text{s}}{2^{l-1}} \!\times\! \frac{W_\text{s}}{2^{l-1}} \!\times\! C_l$ and $\frac{H_\text{t}}{2^{l-1}} \!\times\! \frac{W_\text{t}}{2^{l-1}} \!\times\! C_l$, respectively. }
{The corresponding grids of feature locations at scale $l$, $\Omega_\text{s}^l$ and $\Omega_\text{t}^l$, are of size $\frac{H_\text{s}}{2^{l-1}} \!\times\! \frac{W_\text{s}}{2^{l-1}}$ and $\frac{H_\text{t}}{2^{l-1}} \!\times\! \frac{W_\text{t}}{2^{l-1}}$. The objective of DIM is to compute correspondences between fine-scale source locations $\{\vp_{\text{s},i}^1\}_{i=1,...,|\Omega_\text{s}^1|}$ and fine-scale target locations $\{\vp_{\text{t},i}^1\}_{i=1,...,|\Omega_\text{t}^1|}$, where $\vp$ is 
a 2D vector of integers.}

{To address the computational challenges of DIM, the task can be formulated as a series of coarse-to-fine classification steps. At the coarsest scale $L$, dense correspondence maps $\mC_{\vp_{\text{s},i}^L}^L=\text{softmax}(\mF_\text{s}^L(\vp_{\text{s},i}^L) \odot \mF_\text{t}^L)$ are computed over the entire target grid $\Omega_\text{t}^L$ for each source location $\vp_{\text{s},i}^L=\nearest{\frac{\vp_{\text{s},i}^1}{2^{L-1}}} \in \Omega_{\text{s}}^L$}, where $\odot$ denotes the inner product.

{The process then iterates to finer scales. For each source location $\vp_{\text{s},i}^{l}=\nearest{\frac{\vp_{\text{s},i}^1}{2^{l-1}}} \in \Omega_{\text{s}}^{l}$, the coarser correspondence map $\mC_{\vp_{\text{s},i}^{l+1}}^{l+1}$ is used to determine a sparse search region $\Omega_{\text{t},i}^{l} \subset \Omega_\text{t}^{l}$ (see Sec.~\ref{sec:beam}). Within this region, a sparse correspondence map is computed as:  \vspace{-0.1cm}
\begin{equation} 
\mC_{\vp_{\text{s},i}^{l}}^{l}=\text{softmax}(\mF_\text{s}^{l}(\vp_{\text{s},i}^{l}) \odot \mF_\text{t}^{l}(\Omega_{\text{t},i}^{l})). 
\label{eq:sparse_map}
\end{equation}
This refinement continues down to the finest scale ($l\text{=}1$). The final sparse correspondence map $\mC_{\vp_{\text{s},i}^1}^1$ is computed, and the estimated correspondent $\vp_{\text{t},i}^1$ of $\vp_{\text{s},i}^1$ is defined as the expectation of $\mC_{\vp_{\text{s},i}^1}^1$.}

\subsection{Beam Search}\label{sec:beam}

{A key step in the previously described framework is the definition of the sparse search region $\Omega_{\text{t},i}^l$ at each scale. A simple strategy is to take the argmax (top 1) of $\mC_{\vp_{\text{s},i}^{l+1}}^{l+1}$ with a small local window around it~\cite{tan2022eco,cao2023improving}, but this strategy would precisely fail when the correspondents are widely spread (see Fig.~\ref{fig:teaser}).
}
{Instead, we consider the top $K_{l+1}$ locations of $\mC_{\vp_{\text{s},i}^{l+1}}^{l+1}$ as the set of 2D locations $\Omega_{\text{t},i}^{l}$. Such a strategy is called a beam search~\cite{furcy2005limited}. More precisely, each location $\vp$ from the top $K_{l+1}$ locations is transformed into four locations: $(2\vp\text{+}[0\,0]\tr,2\vp\text{+}[1\,0]\tr,2\vp\text{+}[0\,1]\tr,2\vp\text{+}[1\,1]\tr)$. Thus, in practice, a sparse correspondence map at scale $l$ is evaluated at $4K_{l+1}$ locations.}

{Compared to state-of-the-art dense matching methods that predict a single correspondent hypothesis per source location at each scale, our beam search strategy
ensures that multiple hypotheses are preserved and propagated across scales, addressing ambiguities effectively.}

{To make sure this beam search strategy is computationally feasible, we conducted a multiple hypothesis analysis (see Fig.~\ref{fig:exp_hypo_MD}). We took 10,000 training image pairs from the MegaDepth dataset~\cite{li2018megadepth} with Ground Truth (GT) correspondences $\{\left(\vp_{\text{s},k}^{\text{GT},1}, \vp_{\text{t},k}^{\text{GT},1}\right)\}$ computed using the available depth maps and camera poses. For each scale $l=2,\dots,5$, the GT correspondences are down-sampled as: 
\begin{equation}
\vp_{\text{s},k}^{\text{GT},l} = \nearest{\frac{\vp_{\text{s},k}^{\text{GT},1}}{2^{l-1}}}, \quad 
\vp_{\text{t},k}^{\text{GT},l} = \nearest{\frac{\vp_{\text{t},k}^{\text{GT},1}}{2^{l-1}}}.
\end{equation}
We experimentally observe that it is possible to properly establish $99\%$ of the correspondences with $K_2=4$, $K_3=9$, $K_4=10$ and $K_5=12$, which is indicates that the beam search can be a very efficient strategy. In practice, as we are not provided with \emph{perfect} features, higher values are required. Through experiments, we found $K_2=8, K_3=16, K_4=24, K_5=32$ to be a good trade-off between memory requirements and the capacity to correctly establish correspondences (see Ablation study Fig.~\ref{fig:ablation}).}

\begin{table*}[ht!]
\centering
\small
\caption{{\textbf{Matching accuracy at 3 pixels on MegaDepth-1500~\cite{sarlin2020superglue}, MegaDepth-8-scenes~\cite{edstedt2023dkm} and HPatches~\cite{balntas2017hpatches} for increasing \emph{spread} levels ($\eta$ in pixels).} BEAMER consistently outperforms the state-of-the-art methods: the more the spreading increases, the greater the gap between BEAMER and state-of-the-art methods grows. The gap is smaller on HPatches as the image pairs are less challenging (planar scenes).}}
\vspace{-0.25cm}
\begin{tabular}{@{\hspace{1mm}}c@{\hspace{2mm}}l@{\hspace{3mm}}c
@{\hspace{1mm}}c@{\hspace{1mm}}c@{\hspace{1mm}}c@{\hspace{1mm}}c@{\hspace{3mm}}c@{\hspace{1mm}}c@{\hspace{1mm}}c@{\hspace{1mm}}c@{\hspace{1mm}}c@{\hspace{3mm}}c@{\hspace{1mm}}c@{\hspace{1mm}}c@{\hspace{1mm}}c@{\hspace{1mm}}c@{\hspace{1mm}}}
\specialrule{.1em}{1em}{0em}\\[-1.5ex] 

&\multirow{3}{*}{Method}  
&\multicolumn{14}{@{\hspace{0mm}}c@{\hspace{0mm}}}{Matching Accuracy @3pix (\%) $\uparrow$} \\[1.25ex]
\cline{3-16}
\\[-2ex]
&& \multicolumn{4}{@{\hspace{0mm}}c@{\hspace{0mm}}}{\textbf{MegaDepth-8scenes}} &&  
 \multicolumn{4}{@{\hspace{0mm}}c@{\hspace{0mm}}}{\textbf{HPatches}} &&
 \multicolumn{4}{@{\hspace{0mm}}c@{\hspace{0mm}}}{\textbf{MegaDepth-1500}} \\[0.5ex]
 \cline{3-6} \cline{8-11} \cline{13-16}
\\[-2.ex]
                        && {$\eta$}$\in$[20,40] & [40,60] & [60,80]& [80,100] && {$\eta$}$\in$[20,40] & [40,60] & [60,80]& [80,100] &&  {$\eta$}$\in$[20,40] & [40,60] & [60,80]& [80,100] \\ [.5ex]
         \hline \\[-1.75ex]

& EcoTR~\cite{tan2022eco} \tiny{ECCV'22} & \hspace{3mm} 90.4 & 68.6 & 53.1 & 43.1 &&
\hspace{3mm} 63.0 & 51.1 & 28.3 & 12.4 &&
\hspace{3mm} 85.9 & 68.3 & 54.8 & 48.3 \\

& CasMTR~\cite{cao2023improving} \tiny{ICCV'23} & \hspace{3mm} 92.4 & 71.0 & 57.8 & 47.4 &&
\hspace{3mm} 69.9 & 57.6 & 49.5 & 35.2 &&
\hspace{3mm} 92.5 & 72.7 & 56.8 & 51.0 \\

& DKM~\cite{edstedt2023dkm} \tiny{CVPR'23} & \hspace{3mm} 93.0 & 71.7 & 54.9 & 45.4 &&
\hspace{3mm} 70.9 & 58.4 & 50.4 & 37.4 &&
\hspace{3mm} 93.4 & 75.3 & 60.2 & 53.9 \\

& RoMa~\cite{edstedt2024roma} \tiny{CVPR'24} & \hspace{3mm} \textbf{95.3} & 77.5 & 60.8 & 51.2 &&
\hspace{3mm} 72.8 & 61.3 & 56.8 & 40.1 &&
\hspace{3mm} 94.5 & 78.3 & 64.8 & 59.3 \\

\arrayrulecolor{lightgray}\midrule\arrayrulecolor{black}

& \textbf{BEAMER} & \hspace{3mm} 94.9 & \textbf{83.5} & \textbf{77.2} & \textbf{69.2} &&
\hspace{3mm} \textbf{73.9} & \textbf{63.5} & \textbf{59.7} & \textbf{45.2} &&
\hspace{3mm} \textbf{94.7} & \textbf{83.3} & \textbf{77.6} & \textbf{72.2} \\[-2ex] 

\specialrule{.1em}{1em}{0em}\\[-1.75ex]

\end{tabular}
\label{table:quantitative}
 \end{table*}

\begin{figure*}
    \centering
    \includegraphics[width=0.99\linewidth]{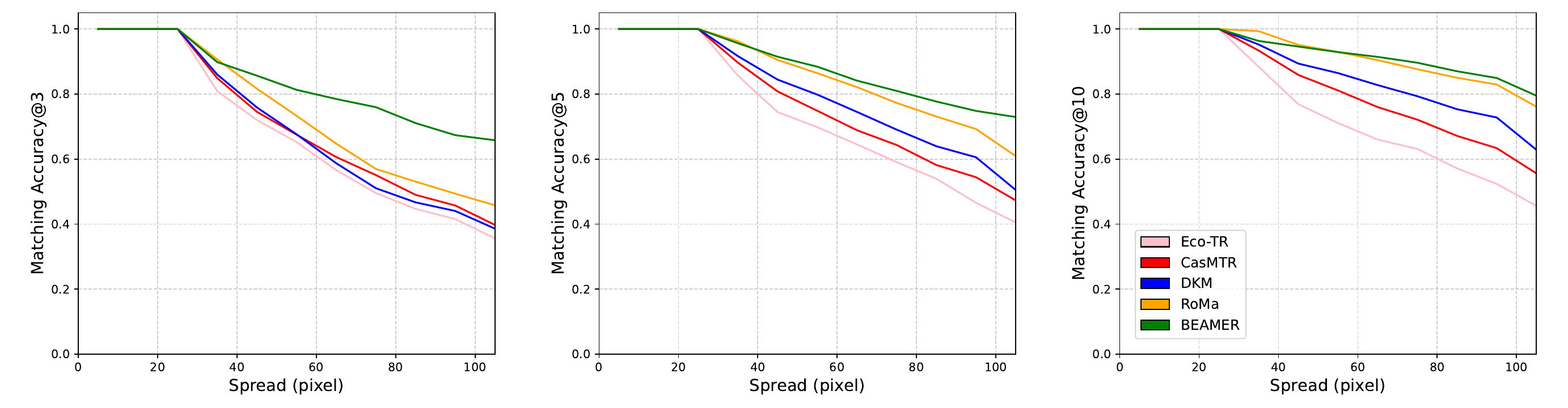}
    \vspace{-4mm}
    \caption{{\textbf{Matching accuracy at 3, 5, and 10 pixels on MegaDepth-8-scenes for increasing \emph{spread} levels.} State-of-the-art methods (EcoTR~\cite{tan2022eco}, CasMTR~\cite{cao2023improving}, DKM~\cite{edstedt2023dkm} and RoMa~\cite{edstedt2024roma}) degrade significantly when the \emph{spread} increases while BEAMER remains robust by effectively preserving and propagating multiple hypotheses.}}
    \label{fig:mm_analysis}
\end{figure*}

\subsection{Network Architecture}\label{sec:architecture}

{Our BEAMER architecture is built on a Siamese Feature Pyramid Network with a ResNet-18 backbone. This backbone extracts multi-scale dense feature maps for 
the source and target images, denoted as $\curl{\mF_\text{s}^l}_{l=1...5}$ and $\curl{\mF_\text{t}^l}_{l=1...5}$, which serve as the input for the beam search mechanism.}

\noindent\textbf{Attention Mechanisms - } {At the coarsest scale ($l=5$), BEAMER employs a dense attention module composed of vanilla self- and cross-attention layers. These layers enable the coarse-scale source and target features to communicate and adapt to each other \cite{sarlin2020superglue}, laying the foundation for accurate correspondence estimation. However, applying dense attention at finer scales is computationally prohibitive due to the resolution of the feature maps.} \\
\noindent {To address this problem, at each refinement scale ($l<5$), for each source location $\vp_{\text{s},i}^{l}$, BEAMER leverages the sparse search region $\Omega_{\text{t},i}^{l}$  (\emph{i.e.} the top $K_{l+1}$ locations identified by the beam search at scale $l+1$) to perform \emph{sparse cross-attention}. For each feature $\mF_\text{s}^{l}(\vp_{\text{s},i}^{l})$ of the source, sparse cross-attention is computed over its corresponding search space $\mF_\text{t}^{l}(\Omega_{\text{t},i}^{l})$ as determined by the beam search, allowing the network to effectively mitigate ambiguities caused by multiple hypotheses. In addition, BEAMER incorporates sparse self-attention layers, which applies the same principle of beam search within the source features and within the target features independently. This process, independent of the correspondence search, enables efficient self-attention while preserving computational feasibility. Together, these operations form a \emph{beam attention module}, which consists of two sparse self-attention layers and two sparse cross-attention layers, efficiently capturing both local and global information at every scale. Technical details are provided in supplementary material.}

\noindent\textbf{Overall Architecture - } {The BEAMER architecture (see Fig.~\ref{fig:architecture}) combines dense and beam attention modules to efficiently preserve and propagate multiple hypotheses across scales. At the coarsest scale ($l=5$), 4 dense attention modules are applied. For the refinement scales ($l=4$ to $l=1$), BEAMER uses $[2, 2, 1, 1]$ beam attention modules, respectively. At each refinement scale, sparse correspondence maps are computed after the beam attention modules. All operations are performed bidirectionally, \emph{i.e.} source $\rightarrow$ target and target $\rightarrow$ source, to predict correspondents for all pixels in both the source and target images.
This design ensures that BEAMER leverages the strengths of dense attention at the coarse scale while efficiently resolving ambiguities at finer scales through beam attention.}

\subsection{Training}\label{sec:training}
{At training-time, we are provided with image pairs and GT correspondences. For each source/target image pair, a set of GT correspondences $\curl{(\vp_{\text{s},k}^{\text{GT},1},\vp_{\text{t},k}^{\text{GT},1})}_{k=1...N}$  is available.
Our objective is to maximize the likelihood of each correspondence at each scale $l=1...L$ to learn to preserve and propagate multiple hypotheses across scales.
In our classification framework, this is equivalent to minimizing 
a 
sum of negative log-likelihood (\emph{a.k.a.} cross-entropy) terms: \vspace{-0.2cm}
\begin{equation}
	 \sum_{k=1}^N \mathcal{L}\parr{\vp_{\text{s},k}^{\text{GT},1},\vp_{\text{t},k}^{\text{GT},1}},\label{eq:losstot}\vspace{-0.2cm}
\end{equation}
where
\begin{equation}
\mathcal{L}\parr{\vp_{\text{s},k}^{1},\vp_{\text{t},k}^{1}} = - \sum_{l=1}^{L} \ln\left(\mC_{\vp_{\text{s},k}^{l}}^{l}\parr{\vp_{\text{t},k}^{l}}\right), \label{eq:loss} \vspace{-0.2cm}
\end{equation}
\noindent with $\vp_{\text{s},k}^{l}=\nearest{\frac{\vp_{\text{s},k}^{1}}{2^{l-1}}}$ and $\vp_{\text{t},k}^{l}=\nearest{\frac{\vp_{\text{t},k}^{1}}{2^{l-1}}}$.
}

\section{Experiments}
\label{sec:experiments}

\subsection{Datasets and Evaluation Protocol}
{We are interested in evaluating the performance of BEAMER and state-of-the-art methods for several increasing spread levels of the neighboring source locations correspondents. To do so, we consider dense GT correspondences from two datasets. The MegaDepth~\cite{li2018megadepth} dataset includes two test sets: the traditional {MegaDepth-1500}~\cite{sarlin2020superglue} consisting of 1500 pairs from two scenes, and {MegaDepth-8scenes}~\cite{edstedt2023dkm}, which offers more diversity with 2400 pairs spanning eight different scenes. The {HPatches}~\cite{balntas2017hpatches} dataset consists of multiple pairs of planar scenes.}

{Given a pair of source/target images, for each source pixel, we consider a patch of 16x16 pixels (\emph{i.e.} the area covered by the corresponding pixel at scale 1/16) and compute the bounding box of their GT correspondents in the target image. The largest side of the bounding box is called the \emph{spread} (in pixels). Doing so allows us to classify the GT correspondences into increasing {spread} levels. The evaluation is performed using the matching accuracy \cite{vilain2024semi} at three pixel error thresholds: 3, 5, and 10 pixels.}
{We compare our method against state-of-the-art dense matching techniques that follow different refinement strategies. {EcoTR}~\cite{tan2022eco} performs coarse-to-fine matching using a sequence of zooms with a local window. {CasMTR}~\cite{cao2023improving} follows a similar coarse-to-fine approach but applies local window refinement at each scale. {DKM}~\cite{edstedt2023dkm} and {RoMa}~\cite{edstedt2024roma} use regression-based strategies for coarse-to-fine matching. Compared to these methods, BEAMER is the only method that seeks to preserve and propagate multiple hypotheses across scales.}

\begin{figure*}[!ht]
    \centering
    \includegraphics[width=0.45\linewidth]{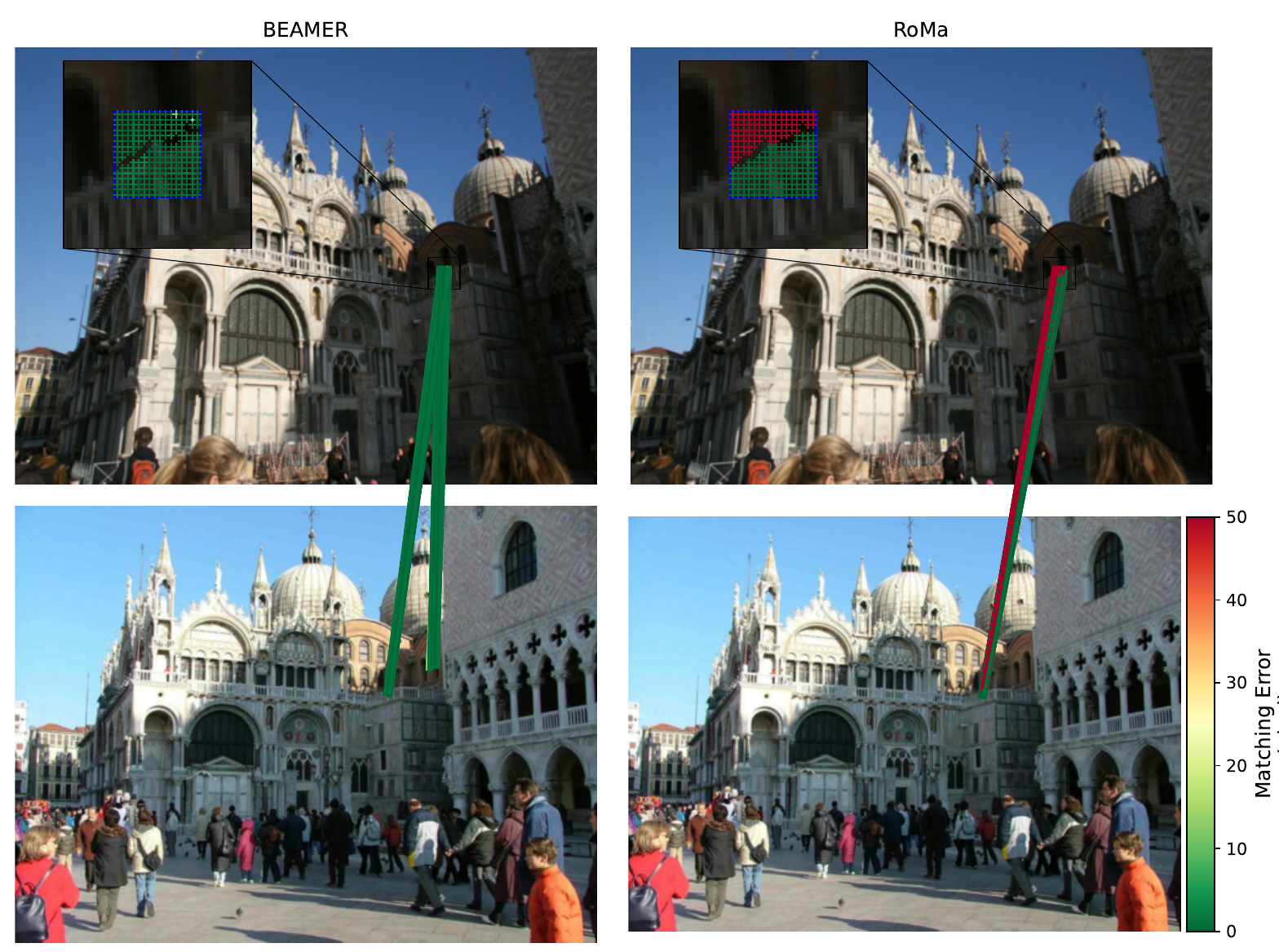} \hspace{0.3cm}
    \includegraphics[width=0.45\linewidth]{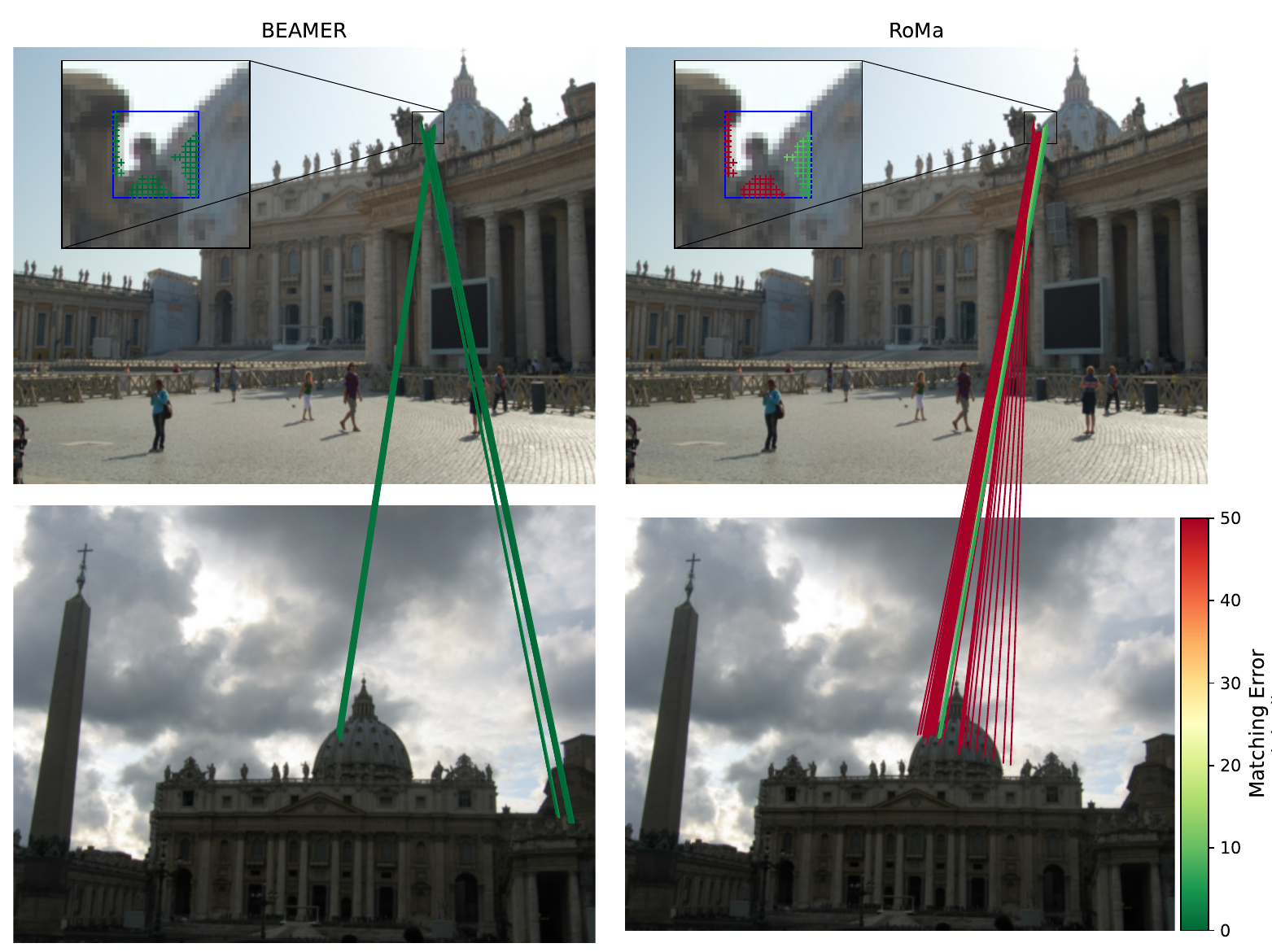}
    \vspace{-2mm}
    \caption{{\textbf{Qualitative comparison: we show the correspondents found by BEAMER and RoMa~\cite{edstedt2024roma} for a $16{\times}16$ source patch.} In these examples, the GT correspondents are located on two different modes. Only correspondences with ground truth are displayed and the line color 
    indicates the matching error in pixels. RoMa, which cannot propagate multiple hypotheses across scales, has difficulty finding correspondents, while BEAMER, designed to preserve and propagate multiple hypotheses across scales, successfully identifies the correspondents.
    }}
    \label{fig:qualitatif}
\end{figure*}

\subsection{Experimental Results}
{Figure~\ref{fig:mm_analysis} shows the matching accuracy at 3, 5, and 10 pixels for several increasing spread levels. We observe that state-of-the-art methods EcoTR~\cite{tan2022eco}, CasMTR~\cite{cao2023improving}, DKM~\cite{edstedt2023dkm} and RoMa~\cite{edstedt2024roma} degrade significantly in accuracy as the spread increases. In contrast, BEAMER’s ability to preserve and propagate multiple hypotheses leads to superior accuracy across all settings.}

{This behavior is illustrated in Fig.~\ref{fig:qualitatif}, where we show the correspondents found by BEAMER and RoMa~\cite{edstedt2024roma} for a $16\times16$ source patch. 
In these examples, the GT correspondents are located on two different modes. RoMa, that is not able to propagate multiple hypotheses across scales,  struggles to find the correspondents, whereas BEAMER that was designed to preserve and propagate multiple hypotheses across scales successfully finds the correspondents. More quantitative results can be found in supp. mat.}

{The quantitative results are summarized in Tab.~\ref{table:quantitative}. BEAMER consistently surpasses state-of-the-art methods: as the spreading increases, the difference between BEAMER and the other methods becomes more pronounced.}

\begin{figure}
    \centering
    \includegraphics[width=0.8\linewidth]{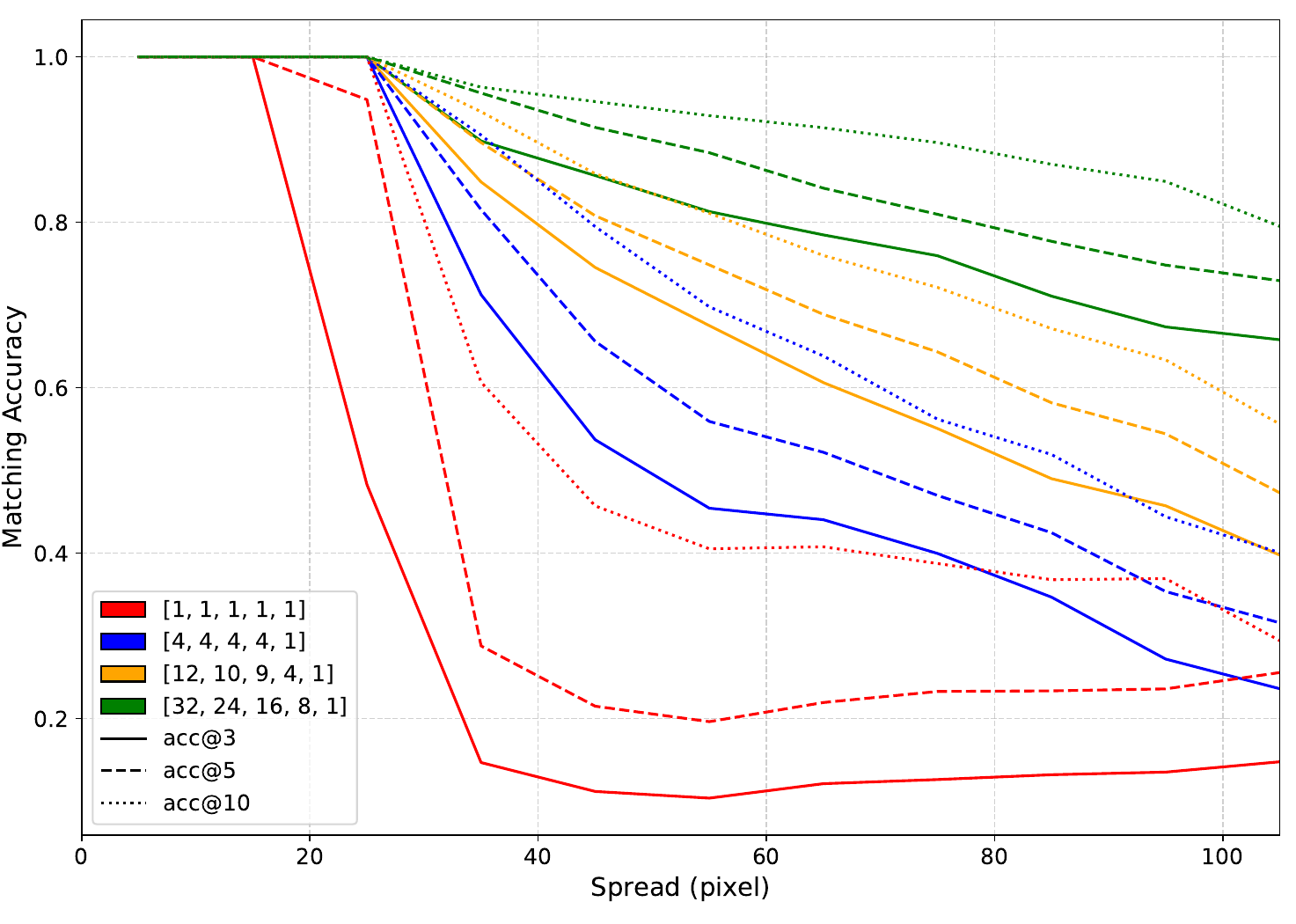}
    \vspace{-4mm}
    \caption{\textbf{Ablation study of the number of hypotheses propagated at each scale [$K_5$,$K_4$,$K_3$,$K_2$] on MegaDepth-8-scenes.} 
    \vspace{-1cm}
    }
    \label{fig:ablation}
\end{figure}

\subsection{Ablation Study}
{Figure~\ref{fig:ablation} presents an ablation study of the number of hypotheses propagated at each scale [$K_5$, $K_4$, $K_3$, $K_2$]. We observe that restricting the search to only the most promising hypothesis [$1$, $1$, $1$, $1$] leads to poor performance. Next, we increase the number of hypotheses to the experimental values from Fig.~\ref{fig:exp_hypo_MD} [$12$, $10$, $9$, $4$], and we can see that the performance consistently improves. Finally, we show that the hyperparameters chosen for BEAMER [$32$, $24$, $16$, $8$] significantly improve the results, compensating for the fact that the feature extractor is not perfect.}

\vspace{-5mm}
\section{Conclusion}

{In this paper, we introduced BEAMER, a novel coarse-to-fine dense image matching approach that learns to preserve and propagate multiple hypotheses across scales using a beam search strategy. Unlike existing coarse-to-fine methods that propagate a single hypothesis per scale, BEAMER effectively maintains candidate correspondents for each source location at each scale, leading to superior robustness in challenging cases such as depth discontinuities and large viewpoint changes. Our results indicate that BEAMER excels in regions where the correspondents of neighboring source locations are widely spread. Future work could explore incorporating more expressive backbones, such as DINOv2, and optimizing memory efficiency to scale BEAMER to even higher resolutions. \\
\textbf{Acknowledgment -- } This project has received funding from the french ministère de l’Enseignement supérieur, de la Recherche et de l’Innovation. This work was granted access to the HPC resources of IDRIS under the allocation 2022-AD011012858 made by GENCI.

\newpage

\section*{Supplementary Material}

\begin{figure*}[!ht]
    \centering
    \begin{minipage}{0.9\textwidth}
        \hspace{0.9cm}
        $l=5$ \hspace{2.5cm} $l=4$ \hspace{2.5cm} $l=3$ \hspace{2.5cm} $l=2$ \hspace{2.5cm} $l=1$
    \end{minipage}
    \includegraphics[width=0.99\linewidth,trim={0 2.5cm 0 2cm},clip]{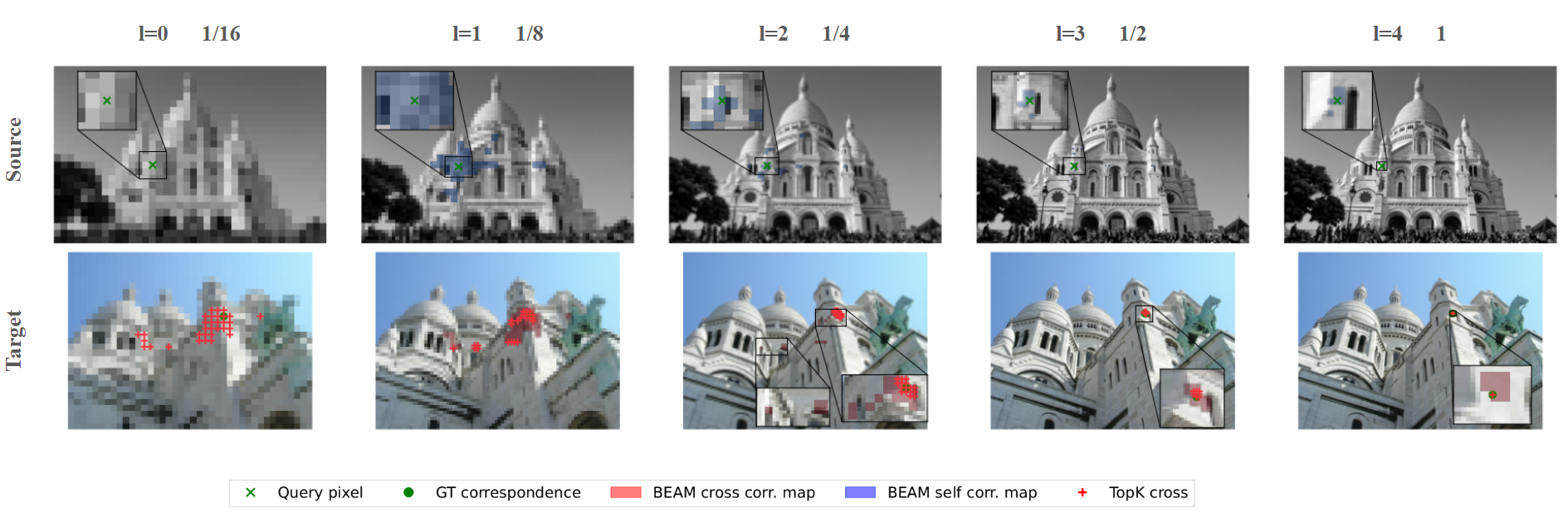}
    \caption{Visualization of the beam search implemented in BEAMER. See the text for details.}
    \label{fig:beam_search_visu}
\end{figure*}

\section{Beam search visualization}

Figure~\ref{fig:beam_search_visu} illustrates the behavior of beam search during the coarse-to-fine matching process.

The source location we consider is in \green{$\times$}. In the target image, the selected hypotheses at each scale are displayed in \red{+}. There are 32 \red{+} at $l=5$ because $K_5=32$, 24 \red{+} at $l=4$ because $K_4=24$, etc. The red areas (of 4 pixels each) correspond to the search regions at each scale (at $l=5$ this area is not represented as it is the whole target image). There are 32 red areas at $l=4$ because $K_{l+1}=K_5=32$. These red areas are the pixels used to perform cross-attention with \green{+}.  At resolutions $l=5$, $4$ and $3$, BEAMER effectively explores distant multiple hypotheses. This ensures that plausible correspondents are considered before progressively refining the search. At finer scales ($l=2,1$), the resolution is sufficiently detailed to focus only on local regions, enabling BEAMER to accurately identify the correct correspondent.

We also display in blue the pixels selected (in the source image) to perform self-attention with \green{+}.
At finer resolutions ($l=1,2$), BEAMER primarily focuses on regions around the query location. However, at coarser resolutions ($l=4,3$), BEAMER also exchanges information with visually similar regions that may introduce ambiguity or regions that may serve as reference points for accurate correspondence estimation.

One important observation from Fig.~\ref{fig:beam_search_visu} is that the red and blue areas represent a significantly smaller subset of the entire pixel grid. This highlights the efficiency of beam search, allowing attention mechanisms to operate effectively even at fine resolutions while limiting the computational cost.

For clarity, we visualize a single correspondence path in Fig.~\ref{fig:beam_search_visu}. However, this process is applied to every pixel in the source image (and every pixel in the target image since BEAMER is bi-directional), ensuring dense matching across the entire image pair.

\section{Implementation Details}
\label{sec:implementation}

\subsection{Backbone Architecture}
The backbone used in BEAMER is a modified version of ResNet18, designed to produce feature maps at every resolutions (1/16, 1/8, 1/4, 1/2, 1). To improve efficiency, we adjust the feature depth at each resolution to the following values: 256 at scale $l=5$ (res. 1/16), 256 at scale $l=4$ (res. 1/8), 128 at scale $l=3$ (res. 1/4), 128 at scale $l=2$ (res. 1/2), and 64 at scale $l=1$ (res. 1).

\subsection{BEAMER Architecture}
For each resolution, different hyperparameters are used in the attention modules. The feature depth varies across scales as in the backbone but is further reduced to reduce the memory footprint: 256 channels at scale $l=5$, 128 channels at scales $l=4$ and $l=3$, 64 channels at scale $l=2$, and 32 channels at scale $l=1$. The number of attention heads and their respective sizes are also adapted (self-attention layers and cross-attention layers have the same hyperparameters at each scale): eight heads of size 64 are used at scale $l=5$, while scales $l=4$, $l=3$, and $l=2$ utilize four heads of size 32. At the finest scale, $l=1$, two heads of size 32 are employed. In every attention module, the feedforward network is replaced with a two-layer convolutional network with a kernel size of 3, ensuring better local consistency in the learned representations.

\begin{figure}[h]
    \centering
    \includegraphics[width=0.99\linewidth]{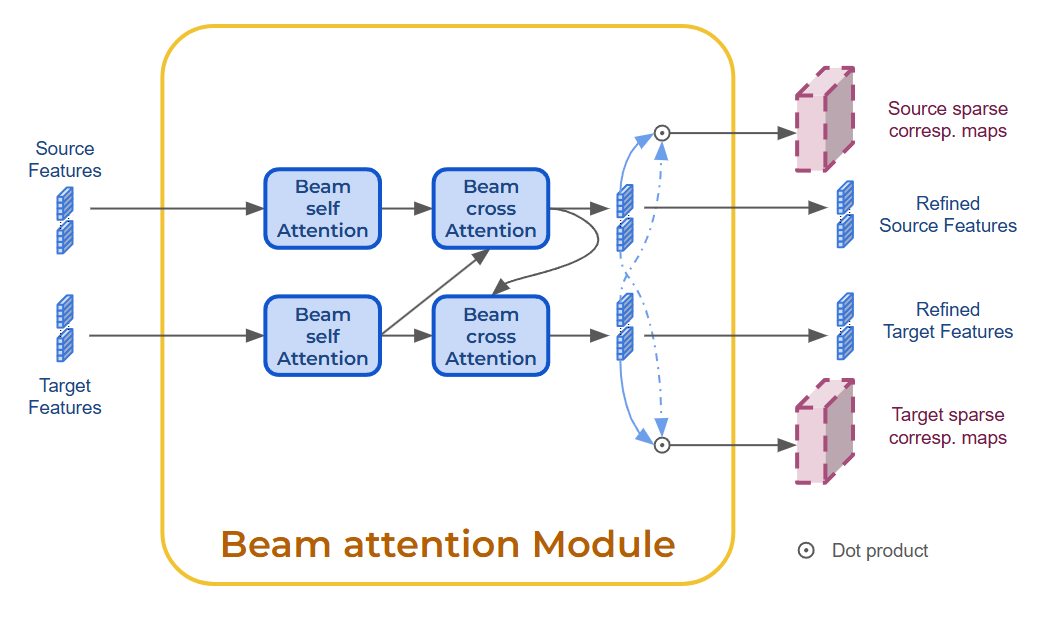}
    \caption{\textbf{Structure of the Beam-attention module.}}
    \label{fig:beam_att_module}
\end{figure}

As described in the main paper, different numbers of attention modules are used at each scale. Specifically, four dense attention modules are employed at the coarsest scale ($l=5$), followed by two beam-attention modules at scales $l=4$ and $l=3$, and one beam-attention module at scales $l=2$ and $l=1$. A more detailed representation of the beam-attention module is provided in Figure~\ref{fig:beam_att_module}, which illustrates its structure and the order of operations.

\subsection{Training Details}
We classically use MegaDepth as training set.
Each training batch consists of a single image pair, where the images are resized such that the largest side is 640 pixels. The training pairs are selected as in DKM, \emph{i.e.} such that half of the image pairs have a minimal overlap of 0.01, while the remaining half contains image pairs with a minimal overlap of 0.35 to include easier cases. The backbone is initially trained from scratch for two hours, only on the coarsest resolution, before integrating it into the full model.

The model is trained using mixed precision (\texttt{FP16}) to optimize computational efficiency. Additionally, gradient checkpointing is employed to further reduce memory consumption at the cost of increased training time. Training is conducted on four Nvidia V100-16GB GPUs, using a dataset consisting of approximately 1.7 million image pairs. The learning rate schedule begins with a warm-up phase of 5000 steps, during which the learning rate is linearly increased from 0 to 0.0001, followed by an exponential decay.

\begin{figure}[h]
    \centering
    \footnotesize
    \begin{tabular}{ccc}
        Loss per resolution & \begin{tabular}{@{}c@{}}\% of GT belonging \\ to corr. maps\end{tabular}  & Matching accuracy \\
    \end{tabular}
    \includegraphics[width=0.32\linewidth, height=0.55\linewidth]{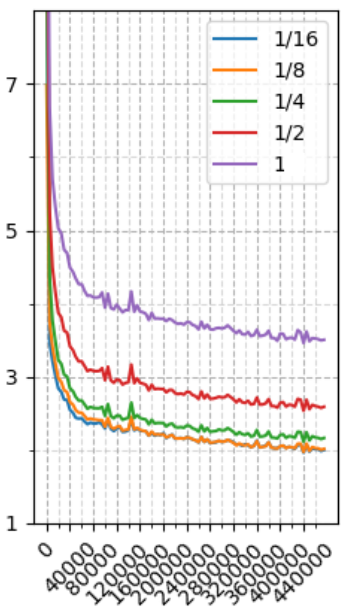}
    \includegraphics[width=0.32\linewidth, height=0.55\linewidth]{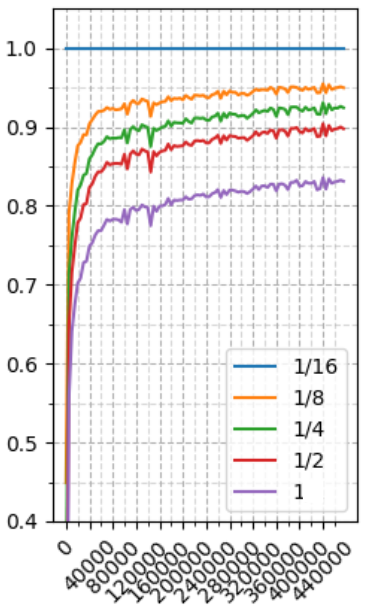}
    \includegraphics[width=0.32\linewidth, height=0.55\linewidth]{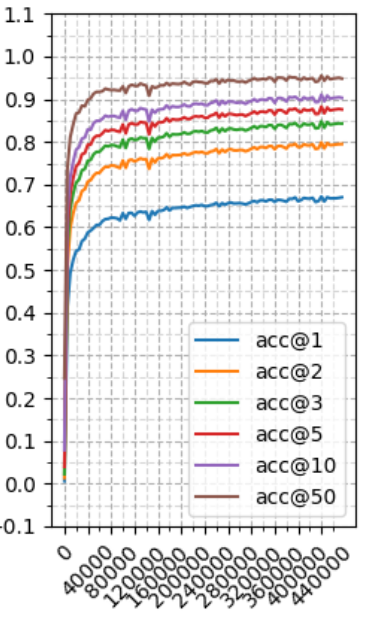}
    \caption{\textbf{Validation metrics during training.}}
    \label{fig:valid_metrics}
\end{figure}

Figure~\ref{fig:valid_metrics} provides an overview of the key validation metrics tracked during training. In addition to monitoring the loss and matching accuracy at each scale, we also report the percentage of ground-truth correspondences that belong to the sparse correspondence maps, which measures the proportion of cases where BEAMER correctly selects the relevant regions to explore during its beam search. The results indicate that BEAMER progressively learns to propagate the relevant hypotheses across scales, achieving a final accuracy of 1 pixel close to 70\%.

\section{Additional qualitative comparison}
Additional qualitative comparisons are shown in Fig.~\ref{fig:supp_qualitative}

\begin{figure*}[h]
    \centering
    \includegraphics[width=0.32\linewidth]{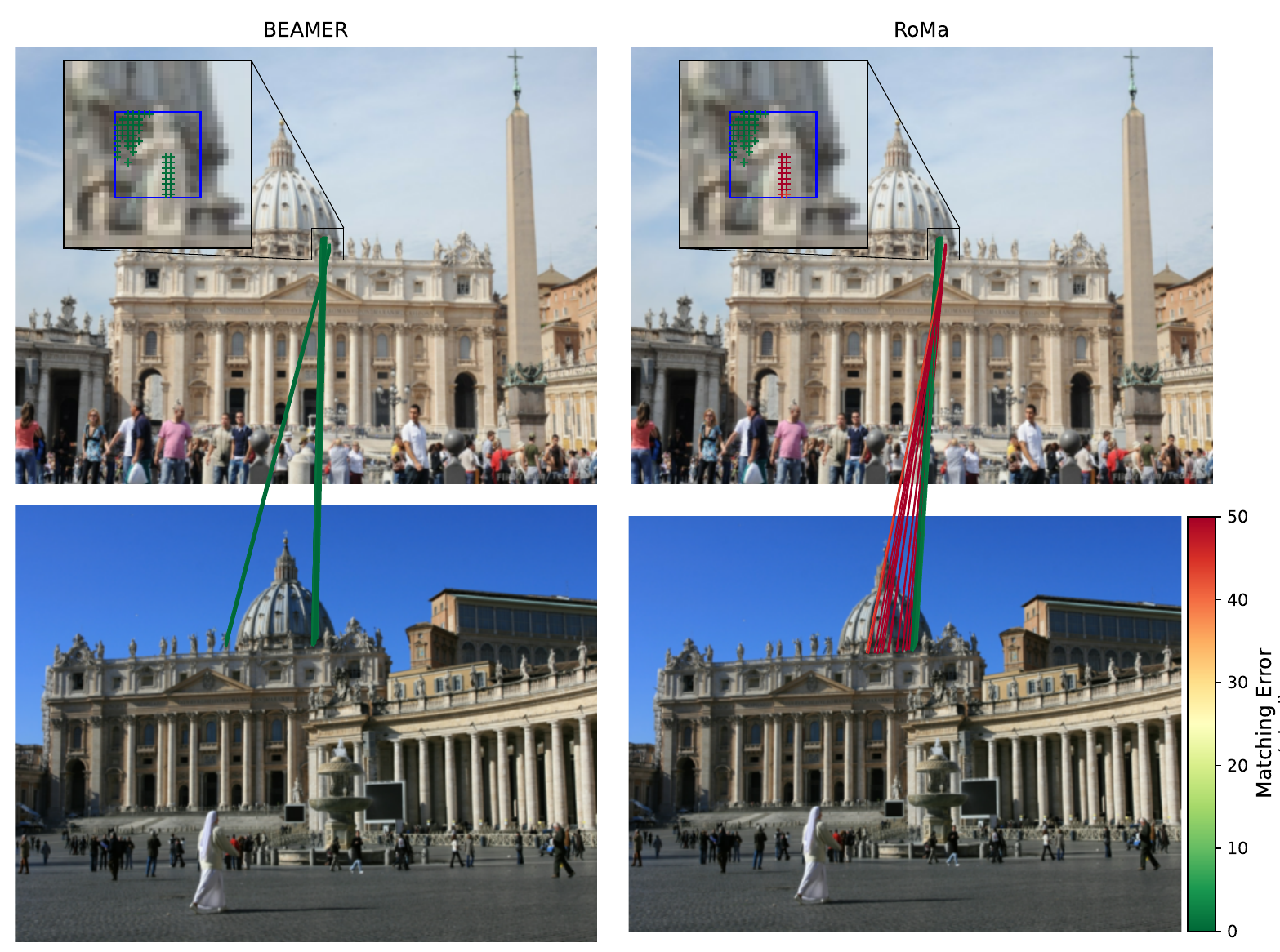}
    \includegraphics[width=0.32\linewidth]{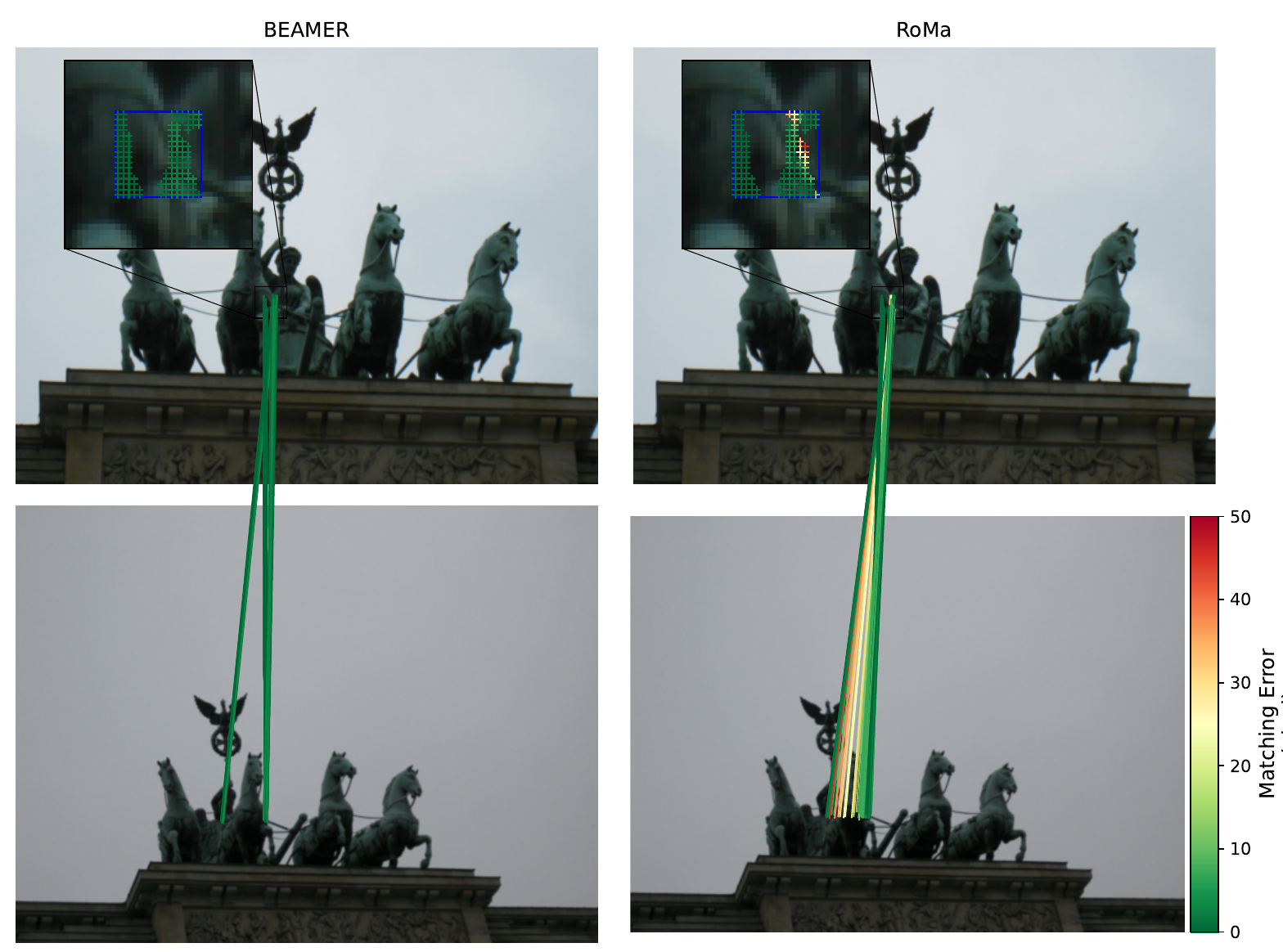} 
    \includegraphics[width=0.32\linewidth]{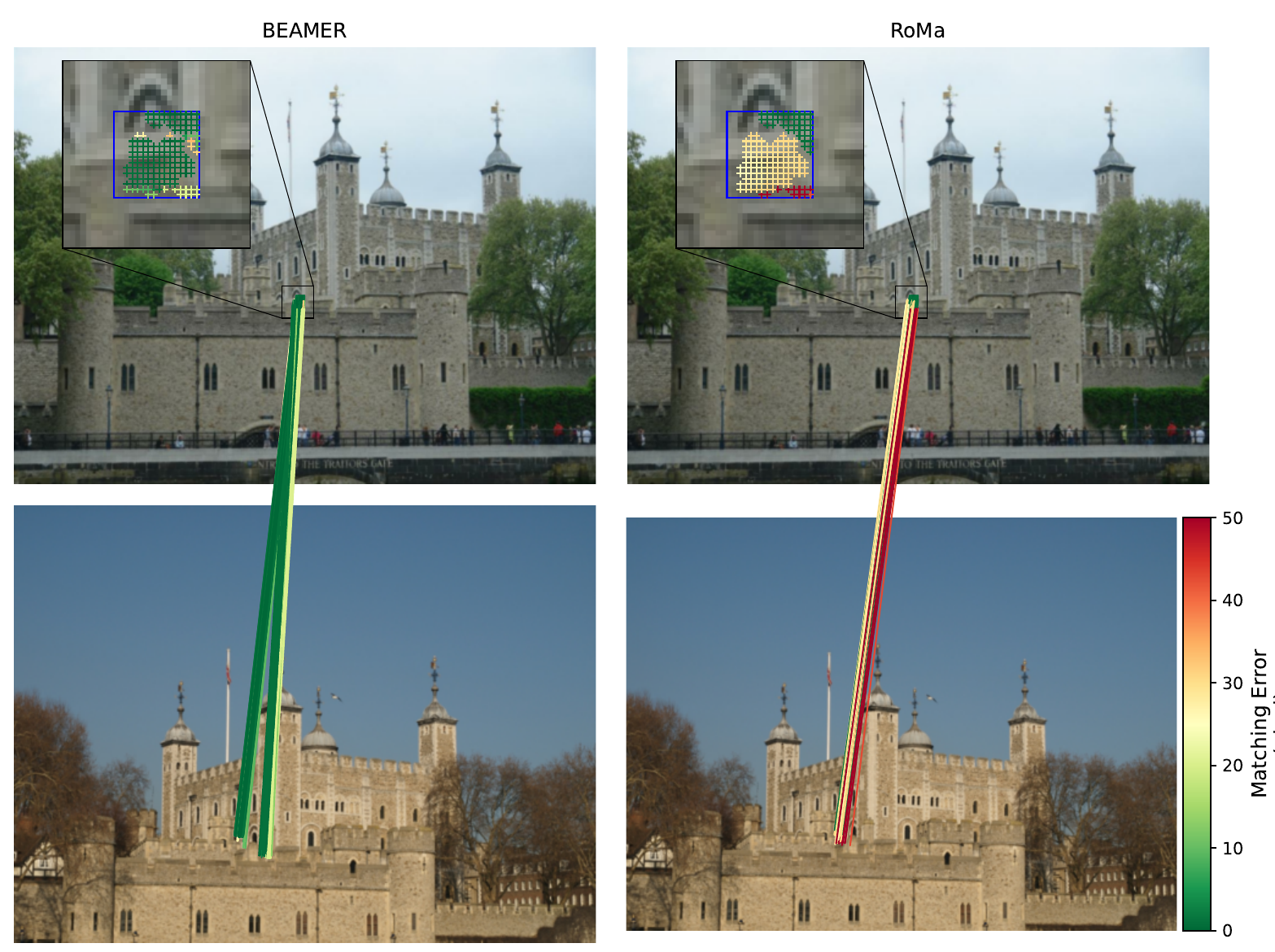} \\
    \vspace{0.5cm}
    \includegraphics[width=0.32\linewidth]{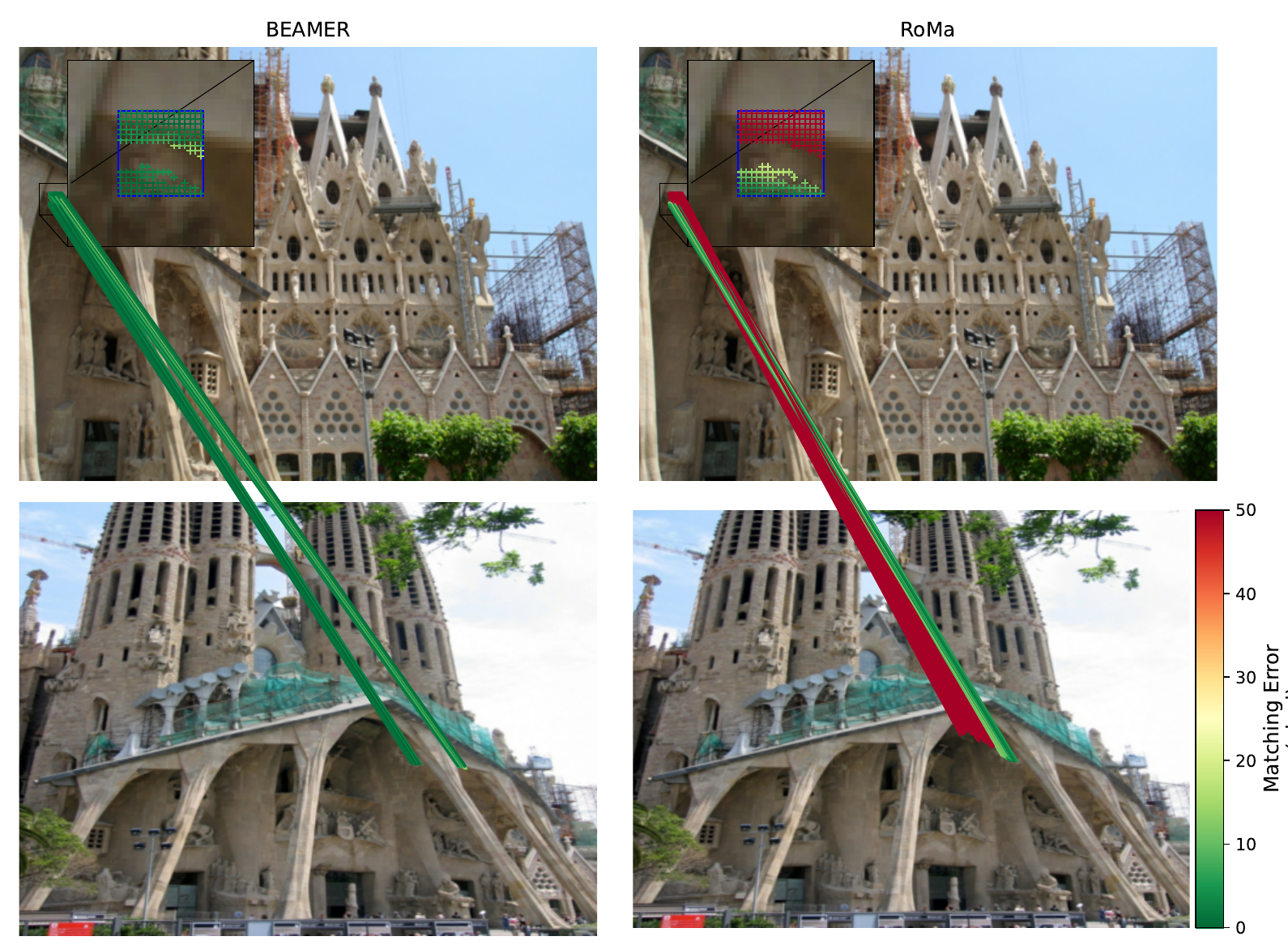}
    \includegraphics[width=0.32\linewidth]{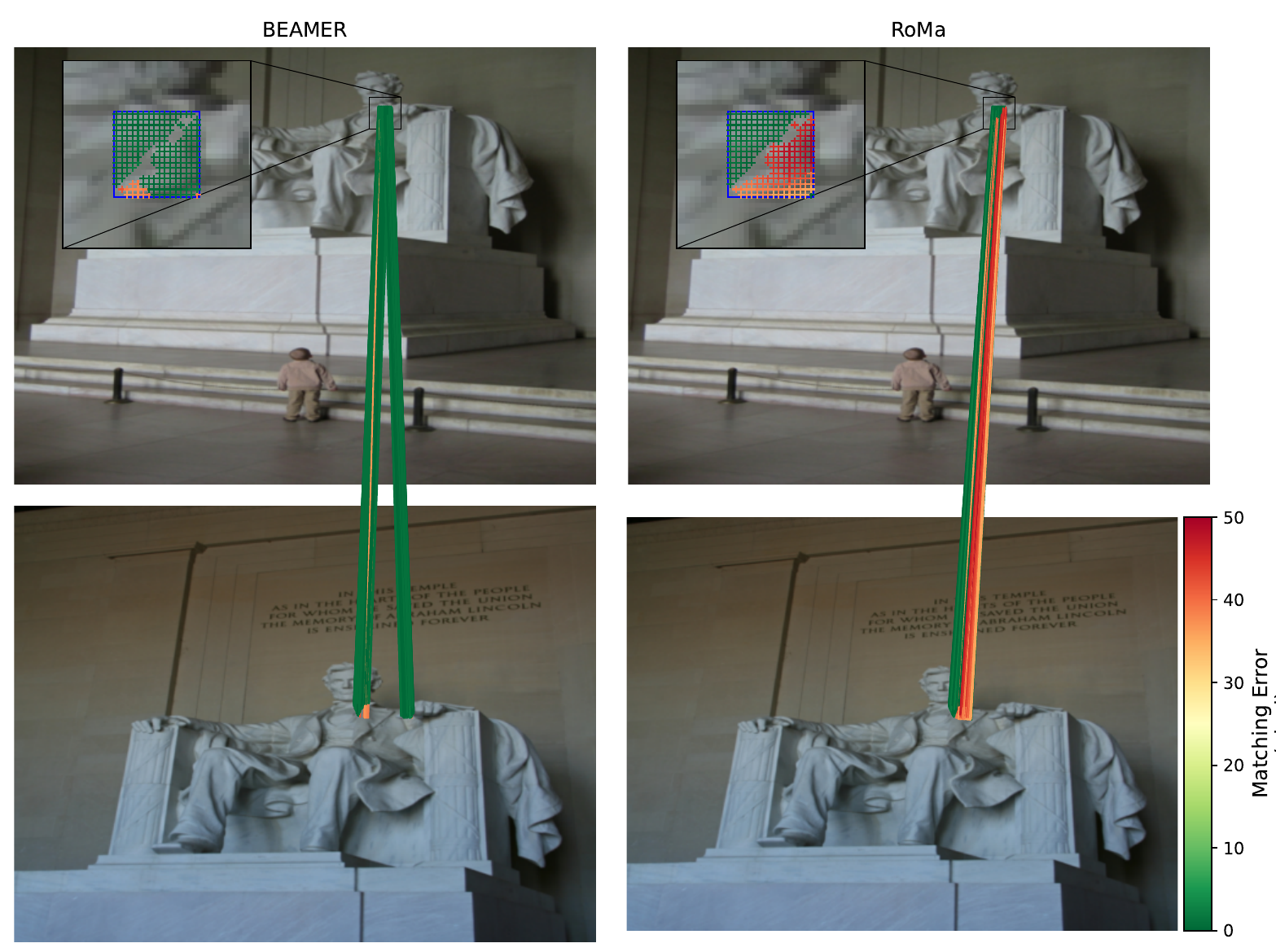}
    \includegraphics[width=0.32\linewidth]{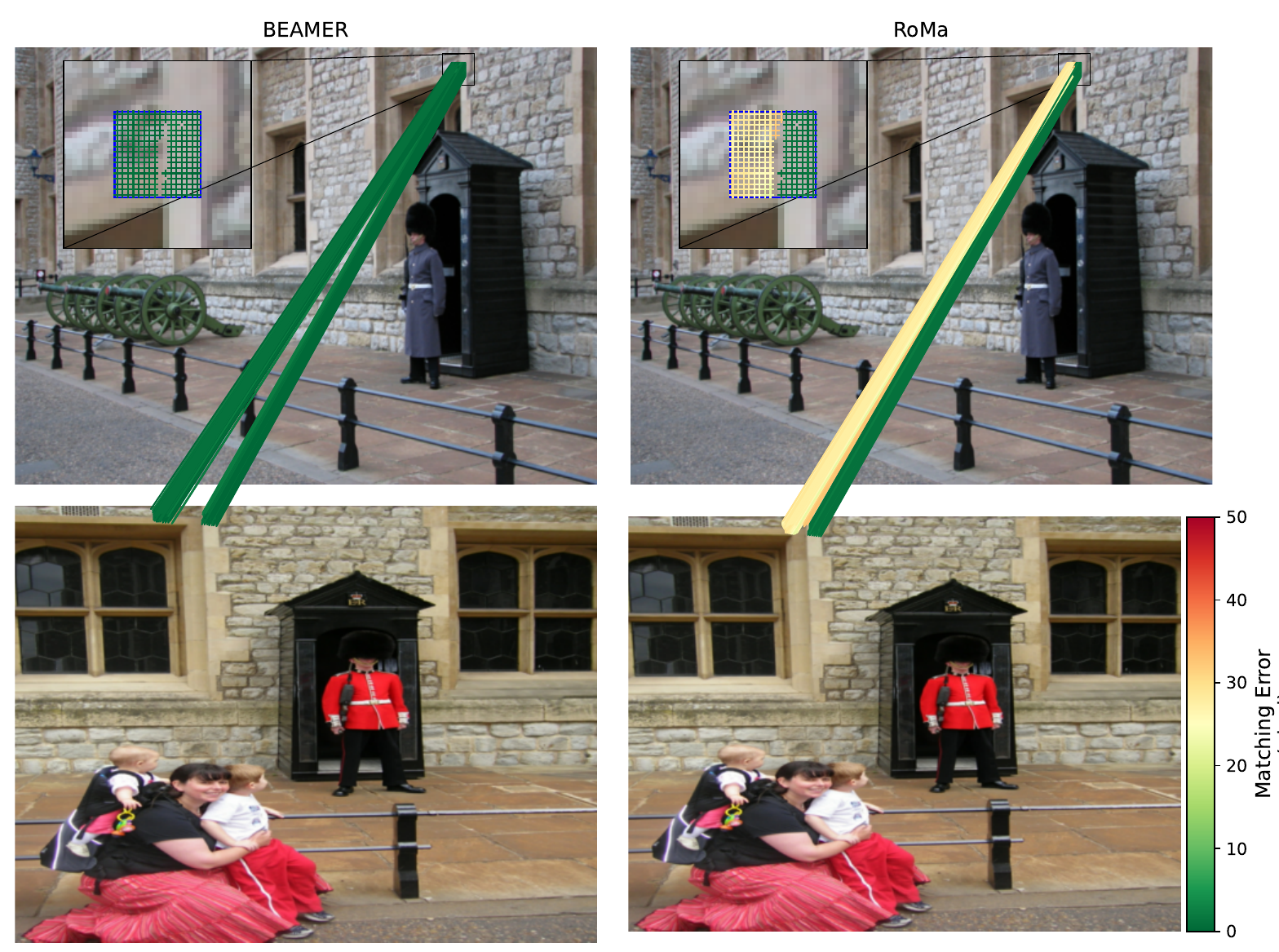}
\caption{{\textbf{Additional qualitative comparison: we show the correspondents found by BEAMER and RoMa for a $16{\times}16$ source patch.} In these examples, the GT correspondents are located on two different modes. Only correspondences with ground truth are displayed and the line color
    indicates the matching error in pixels. RoMa, which cannot propagate multiple hypotheses across scales, has difficulty finding correspondents, while BEAMER, designed to preserve and propagate multiple hypotheses across scales, successfully identifies the correspondents.
    }
}
\label{fig:supp_qualitative}
\end{figure*}

\bibliographystyle{IEEEbib}
\bibliography{strings,refs}

\end{document}